\documentclass[12pt,letterpaper]{article}
\usepackage[T1]{fontenc}
\usepackage[utf8]{inputenc}
\usepackage{newtxtext,newtxmath}
\usepackage[letterpaper,margin=1in,headheight=15pt,headsep=0.25in]{geometry}
\usepackage{setspace,microtype,indentfirst}
\usepackage{booktabs,tabularx,longtable,array}
\usepackage{graphicx,tikz}
\usetikzlibrary{arrows.meta,positioning,fit,calc}
\usepackage{enumitem}
\usepackage[authoryear,round,sort,longnamesfirst]{natbib}
\usepackage{xurl}
\usepackage[hidelinks]{hyperref}
\shortcites{nistplaybook,iso42005,cri2026guide,cri2026reference}
\usepackage{caption}
\usepackage{fancyhdr}
\usepackage{titlesec}
\newcommand{\RaggedRight}{\raggedright\setlength{\parindent}{0.5in}}
\RaggedRight
\setlist{nosep,leftmargin=0.5in}
\newcolumntype{Y}{>{\raggedright\arraybackslash}X}
\newcolumntype{P}[1]{>{\raggedright\arraybackslash}p{#1}}
\newcommand{\colhead}[1]{\centering\arraybackslash #1}

\titleformat{\section}{\normalfont\normalsize\bfseries\centering}{}{0pt}{}
\titleformat{\subsection}{\normalfont\normalsize\bfseries}{}{0pt}{}
\titleformat{\subsubsection}{\normalfont\normalsize\bfseries\itshape}{}{0pt}{}
\titlespacing{\section}{0pt}{0pt}{0pt}
\titlespacing{\subsection}{0pt}{0pt}{0pt}
\titlespacing{\subsubsection}{0pt}{0pt}{0pt}
\fancypagestyle{plain}{\fancyhf{}\fancyhead[L]{AI-GRACE: USE-CASE OPERATIONALIZATION}\fancyhead[R]{\thepage}}
\newcommand{\articletitle}{AI-GRACE: A Use-Case Operationalization Framework for Agentic AI}
\newcommand{\articlesubtitle}{From Organizational Objectives and Obligations to Deployment Capabilities and Architecture}
\newcommand{\apanote}[1]{\par\begingroup\normalsize\doublespacing\RaggedRight\noindent\textit{Note.} #1\par\endgroup}

\newcommand{\grace}{AI-GRACE}
\newcommand{\rail}{RAIL}
\newcommand{\doi}[1]{\href{https://doi.org/#1}{\nolinkurl{https://doi.org/#1}}}
\hypersetup{pdftitle={AI-GRACE: A Use-Case Operationalization Framework for Agentic AI},pdfauthor={John Cuneo, David Chun, Gaurav Khanna},pdfsubject={Conceptual method and illustrative retail banking application}}
\begin{document}
\thispagestyle{fancy}
\vspace*{2\baselineskip}
\begin{center}
\textbf{\articletitle:\\ \articlesubtitle}\par
\vspace{\baselineskip}
John Cuneo\textsuperscript{1}, David Chun\textsuperscript{2}, and Gaurav Khanna\textsuperscript{3}\par
\textsuperscript{1}University of Miami\par
\textsuperscript{2}Columbia University\par
\textsuperscript{3}Stanford University\par
\end{center}
\vfill
\section*{Author Note}
John Cuneo, ORCID: \url{https://orcid.org/0009-0000-5372-2953}; David Chun, ORCID: \url{https://orcid.org/0009-0002-8348-3580}.

The authors' account of the framework's origin and use of AI assistance appears in Appendix~\ref{app:authorship}.

Correspondence concerning this article should be addressed to John Cuneo. Email: \href{mailto:cuneojn@miami.edu}{cuneojn@miami.edu}. Coauthor contacts: David Chun, \href{mailto:dc4029@columbia.edu}{dc4029@columbia.edu}; Gaurav Khanna, \href{mailto:gkhanna@stanford.edu}{gkhanna@stanford.edu}.
\clearpage
\renewenvironment{abstract}{\section*{Abstract}\noindent}{\par}
\begin{abstract}
Organizations deploying agentic artificial intelligence must determine more than whether a model is trustworthy; they must establish what to validate, control, and observe for a use case to deliver its intended outcome while meeting applicable obligations. This paper proposes AI-GRACE (Agentic Intelligence-Governance, Risk, Assurance, Controls, and Evidence) as a use-case operationalization framework connecting organizational governance with technical implementation. The proposal draws on professional observations and a purposive synthesis of standards and literature, using design science to frame the method contribution and situational method engineering to guide contextual tailoring and reuse. The framework establishes objectives and obligations and then assesses risks in seven proposed domains, including mission and value realization. It derives requirements for assurance before deployment, runtime controls, and evidence, which guide capability qualification, gap assessment, and a logical architecture. An Agent Operating Envelope specifies permitted actions and escalation conditions, while Risk-Aligned Independence Levels (\rail{}) summarize the authorized independence. A fictional retail banking application illustrates the method. The contribution is a traceable basis for deciding what an organization must implement, what it already supports, and what remains unresolved. Empirical evaluation must establish whether it improves deployment decisions, efficiency, and reuse.
\end{abstract}
\noindent\hspace*{0.5in}\textit{Keywords:} agentic AI, AI governance, use case operationalization, technical capabilities, reference architecture

\clearpage
\section*{\articletitle: \articlesubtitle}
An organization may want an AI assistant to improve some set of business metrics (e.g., service, complete routine requests, control operating costs). A deployment decision must establish whether the assistant can do the work well, what it can access and change, and whether existing systems can enforce the required boundaries. A risk assessment that does not reach those implementation questions leaves a critical part of the deployment problem unsolved.

Research has long identified the challenge of moving from responsible AI principles to practice \citep{morley2020,papagiannidis2025}. Engineering methods, governance architectures, and delegation specifications provide the foundations for this work \citep{fetzer2026,huang2025,arora2026}. Deployment teams must still select, connect and qualify the capabilities required for their use case.

An agent may retrieve information, select tools, modify records, or coordinate with other agents. As an agent pursues goals with less immediate human involvement, the directness, scale, and duration of its effects can change. Human responsibility remains \citep{chan2023}.

The same large language model can support a read-only assistant or an agent authorized to change operational records. Model evaluation alone cannot determine whether either deployment is justified. This paper asks: \emph{How can an organization translate the objectives, obligations, and risks of an agentic AI use case into qualified technical capabilities, a logical architecture, and an accountable deployment decision?} The expected output is an implementable specification of what the organization needs, what it already has, and what remains to be resolved.

\emph{Agent capabilities} describe what an agent can do. \emph{Organizational enabling capabilities} describe what the organization can provide to deploy, evaluate, constrain, and operate it. AI-GRACE specifies a method for deriving those enabling requirements, qualifying implementations, and retaining a deployment record that supports reassessment and reuse.

RAIL summarizes the independence authorized through that assessment while the operating envelope defines the actual permissions. The objective is risk-adjusted value. A deployment with less independence can be the appropriate outcome when it delivers the intended benefit with lower cost or exposure.

\section{Theoretical Background and Related Work}
\subsection{Organizational Governance and Technical Implementation}
The NIST AI Risk Management Framework organizes AI risk management through GOVERN, MAP, MEASURE, and MANAGE \citep{nist2023}. Its Playbook supplies actions that organizations select according to their circumstances \citep{nistplaybook}.

ISO/IEC 42001 establishes an AI management system, including objectives, risk and impact assessment, and risk treatment with a documented rationale for control selection \citep[clause 6.1.3]{iso42001}. ISO/IEC 42005 addresses impacts on people and society, including thresholds, approvals, and review \citep[clauses 5--6]{iso42005}. ISO/IEC 42001 also addresses resources, architecture documentation, and operational monitoring \citep[Annex B.4 and B.6]{iso42001}.

The Cyber Risk Institute (CRI) FS AI RMF supplies 230 control objectives with implementation guidance and illustrative controls and evidence, while distinguishing its enterprise scope from a prescription for an individual use case (CRI, \citeyear{cri2026guide}, \citeyear{cri2026reference}). AI-GRACE uses these resources to derive deployment requirements within the institution's existing governance arrangements.

\citet{reuel2025} describe technical AI governance as analysis and tools for identifying governance needs, assessing interventions, and enabling enforcement or compliance. AI-GRACE addresses the organizational task of turning a proposed use case into requirements that engineering and operations can implement and assess.

\subsection{Responsible Engineering as a Methodological Foundation}
\citet{fetzer2026} developed MeRGE through action design research with interviews and three realized use cases. Its method spans precheck, conceptualization, development, quality, and deployment, with an architecture containing models, orchestration, data, tools, and risk mitigation. Testing covers functional, security, load, performance, stress, and integration requirements. MeRGE is a methodological predecessor. Its discussion identifies agentic architectures, autonomy, and translation of normative initiatives as further research directions.

AI-GRACE extends this line of work by qualifying the enabling capabilities for an agentic use case and connecting architecture, authority, and evidence to reassessment. Situational method engineering provides a basis for tailoring methods to context, supporting the proposed jurisdictional, sector, and organizational profiles \citep{henderson2010}.

\subsection{Agentic Risk, Control Architectures, and Delegated Authority}
The Agentic Risk \& Capability (ARC) Framework relates components, design, and agent capabilities to contextual risks and technical controls \citep{khoo2025}. NIST AI RMF, OWASP, MITRE ATLAS, and Cisco's Integrated AI Security and Safety Framework can inform the identification of relevant safety and security risks \citep{nist2023,owasp2025,mitreatlas,chang2025}. IMDA's agentic AI framework addresses risk boundaries, human accountability, and technical and nontechnical measures \citep{imda2026}.

\citet{koch2026} connects governance objectives to enforceable controls, architectural placement, ownership, and assurance evidence. AAGATE supplies a NIST-aligned governance platform and control plane \citep{huang2025}. These are close antecedents to AI-GRACE's implementation requirements.

\citet{feng2025} distinguish autonomy levels through the user's role, while \citet{zheng2026} separate autonomous capability from allowed autonomy. \citet{arora2026} propose records for justifying agency and specifying delegated boundaries. RAIL draws on the established distinction between capability and permission to summarize the resulting authorization.

Architecture, controls, autonomy levels, and contextual assessment have these substantive antecedents. AI-GRACE makes capability qualification, service viability, and the completed deployment record central to their application.

\section{Research Approach and Framework Construction}
AI-GRACE is a prescriptive method proposal informed by design science \citep{peffers2007,gregor2013}. Situational method engineering provides a basis for tailoring and reusing assessment activities and records according to the use case context \citep{henderson2010}. AI-GRACE's design proposition is that explicit links among objectives, obligations, risks, capabilities, architecture, and evidence can make deployment decisions more complete and reviewable. The fictional application demonstrates this reasoning.

The authors' professional observations identified a practical deployment problem: organizations needed to translate AI objectives and obligations into technical capabilities. They developed an initial framework, then used a purposive synthesis of governance standards, engineering methods, agentic risk frameworks, and autonomy research to refine its definitions and outputs.

The analysis uses the full texts of ISO/IEC 42001 and 42005, the NIST Playbook, CRI implementation materials, and MeRGE, supplemented by primary publications and official financial services sources. Recent preprints are treated as proposals. The banking example shows how changes in actions and authority affect obligations, capabilities, and architecture.

The seven risk domains are proposed coverage prompts, grouped around concerns that require different questions, evidence, expertise, or treatment. The assessment keeps related causes and consequences together, even when they cross domains.

\section{The AI-GRACE Method}
\subsection{Unit of Assessment, Participants, and Outputs}
AI-GRACE specifies the core method and outputs for assessing an AI use case in its deployment context. The assessment covers the workflow, users, affected parties, agents, models, tools, data, dependencies, and proposed authority. The use case owner and technical architect can lead the work with domain operations, engineering, security, privacy, risk, compliance, and the functions accountable for production services.

A completed assessment explains how the use case's objectives and obligations lead to a deployment decision. It connects intended impacts and risks to required capabilities and a logical architecture, showing what the organization can support and where gaps remain. Appendix~\ref{app:record} lists the minimum record; existing organizational systems can hold its linked parts. The decision specifies an authorized operating envelope and RAIL, or withholds authorization, while recording desired and conditional future states separately.

\begin{figure}[htbp]
\caption{From Use Case Intent to a Qualified Deployment Design}
\label{fig:method}
\raggedright
\begingroup
\singlespacing\hyphenpenalty=10000\exhyphenpenalty=10000
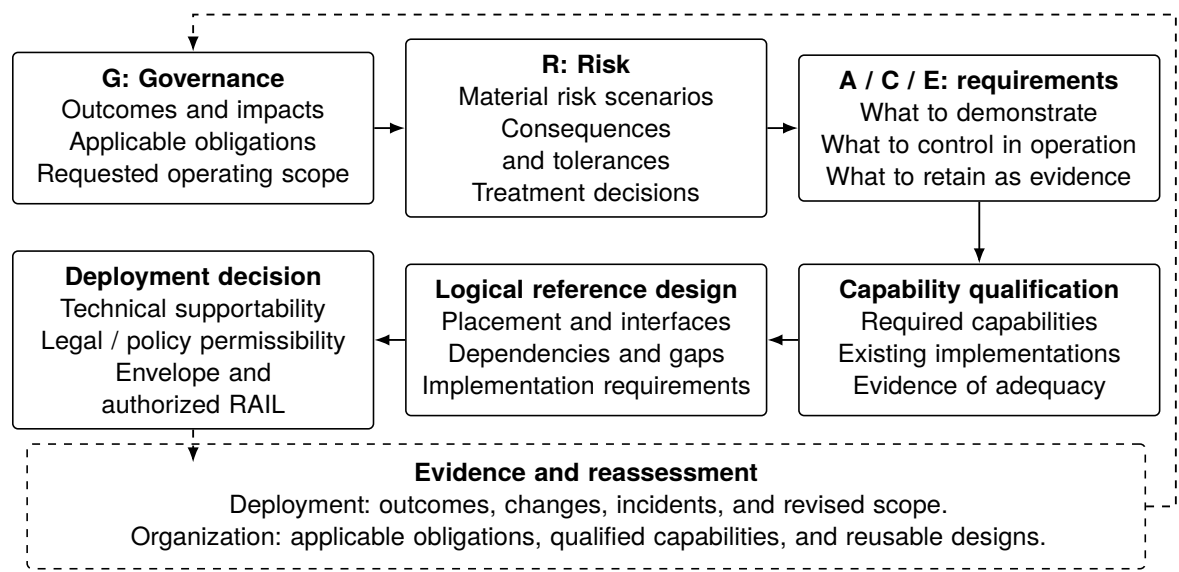
\begin{tikzpicture}[font=\fontsize{10}{12}\selectfont\sffamily,box/.style={draw,line width=0.6pt,rounded corners=2pt,align=center,inner sep=6pt,text width=4.35cm,minimum height=1.8cm},arr/.style={-{Latex[length=2mm]},line width=0.7pt}]
\node[box] (g) at (0,0) {\textbf{G: Governance}\\Outcomes and impacts\\Applicable obligations\\Requested operating scope};
\node[box] (r) at (5.2,0) {\textbf{R: Risk}\\Material risk scenarios\\Consequences and tolerances\\Treatment decisions};
\node[box] (ace) at (10.4,0) {\textbf{A / C / E: requirements}\\What to demonstrate\\What to control in operation\\What to retain as evidence};
\node[box] (fit) at (10.4,-2.8) {\textbf{Capability qualification}\\Required capabilities\\Existing implementations\\Evidence of adequacy};
\node[box] (arch) at (5.2,-2.8) {\textbf{Logical reference design}\\Placement and interfaces\\Dependencies and gaps\\Implementation requirements};
\node[box] (auth) at (0,-2.8) {\textbf{Deployment decision}\\Technical supportability\\Legal / policy permissibility\\Envelope and authorized RAIL};
\node[draw,dashed,line width=0.6pt,rounded corners=2pt,align=center,text width=14.3cm,inner sep=7pt] (feed) at (5.2,-5.0) {\textbf{Evidence and reassessment}\\Deployment: outcomes, changes, incidents, and revised scope.\\Organization: applicable obligations, qualified capabilities, and reusable designs.};
\draw[arr] (g)--(r);\draw[arr] (r)--(ace);\draw[arr] (ace)--(fit);\draw[arr] (fit)--(arch);\draw[arr] (arch)--(auth);
\draw[arr,dashed] (auth.south)--(0,-4.42);
\draw[arr,dashed] (feed.east)--(13.0,-5.0)--(13.0,1.5)--(0,1.5)--(g.north);
\end{tikzpicture}
\endgroup
\apanote{G = Governance; R = Risk; A = Assurance; C = Controls; E = Evidence; RAIL = Risk-Aligned Independence Levels. Dashed arrows show feedback for reassessment and reuse.}
\end{figure}

Figure~\ref{fig:method} shows the reasoning path. Teams revisit earlier decisions when evaluations, architectural constraints, or evidence requirements expose a problem. The method supports iteration within a use case and reuse across later assessments.

\subsection{Governance: Establish the Outcome and Obligations}
Governance establishes why the use case should exist and what success requires. The charter states intended outcomes and measurable acceptance conditions, then identifies affected parties and accountable owners. It sets the boundaries for proposed actions and the use of data and tools, including exclusions (\citealp[clauses 4 and 6.2]{iso42001}; \citealp[MAP 1.1, 1.3--1.4]{nist2023}). It identifies jurisdictions and institutional roles, and a credible alternative such as conventional automation or a service delivered by people.

The obligation record distinguishes law, regulation, contract, professional duties, adopted policy, and strategic commitments. Each entry identifies its source, applicability, owner, and status. Revenue or cost objectives cannot override applicable duties or mandatory legal requirements.

Applicability depends on context: the EU AI Act requires attention to intended purpose and organizational role \citep[Articles 2, 6, 16, and 26]{euaiact2024}. Impact assessment considers foreseeable benefits and harms beyond the sponsoring organization, including clients, employees, and society \citep[clauses 6.7--6.9]{iso42005}.

\subsection{Risk: Identify What Could Defeat the Outcome}
The register describes how each scenario could cause harm or prevent the intended value, identifying causes and the people or assets affected. It links the scenario to the relevant objectives and obligations and assesses likelihood and consequence under stated assumptions. The record assigns a decision owner and explains how existing measures and treatment address the risk, including residual concerns and uncertainty. Organizations can use their established scales; a risk score does not mechanically determine RAIL.

Table~\ref{tab:domains} gives seven proposed coverage domains and selected source anchors. The grouping is a design interpretation, not an endorsed taxonomy.

\begingroup\normalsize\singlespacing
\setlength{\LTleft}{0pt}
\setlength{\LTright}{\fill}
\begin{longtable}{P{3.25cm}P{\dimexpr(\textwidth-6\tabcolsep-3.25cm-3.7cm)/1\relax}P{3.7cm}}
\caption{Risk Domains and Selected Conceptual Anchors}
\label{tab:domains}\\
\toprule \colhead{Domain} & \colhead{Concern} & \colhead{Selected anchors} \\
\midrule\endfirsthead
\toprule \colhead{Domain} & \colhead{Concern} & \colhead{Selected anchors} \\
\midrule\endhead
Mission \& Value & Poor adoption, task failure, delay, resource use, or cost greater than expected benefit. & NIST MAP 1.3--1.4; CRI MP-1.4.1, MS-2.12.3 \\
AI Behavior \& Authority & Without an attacker, the agent reasons poorly, mishandles ambiguous input, selects tools incorrectly, drifts from its goal, or attempts unauthorized action. & NIST MEASURE 2.5; ARC; delegation research \\
Safety \& Human Impact & Harmful reliance, manipulation, unfair treatment, physical or psychological harm, or impaired access affects people or communities. & NIST MEASURE 2.6, 2.11; ISO/IEC 42005 clauses 6.7--6.9; \citealp{chan2023} \\
Cybersecurity & An adversary compromises or misuses the agent, identities, data, tools, dependencies, or connected systems. & NIST MEASURE 2.7; ARC; OWASP \\
Privacy & Personal information is improperly collected, accessed, inferred, retained, used, transferred, or disclosed, with or without an attacker. & NIST MEASURE 2.10; ISO/IEC 42005 \\
Operational Resilience & Availability, latency, capacity, recovery, dependencies, or maintainability cannot sustain the required service. & NIST MEASURE 2.3, 2.7; ISO/IEC 42001 Annex B.6.2.6; MeRGE quality phase \\
Legal, Regulatory \& Policy & The use case or its operation fails to satisfy an applicable obligation. & NIST GOVERN 1.1; ISO/IEC 42001 clause 6.1.3; CRI GV-1.1.1 \\
\bottomrule
\end{longtable}
\apanote{NIST = National Institute of Standards and Technology (AI RMF 1.0, \citeyear{nist2023}); CRI = Cyber Risk Institute (Control Objective Reference Guide, \citeyear{cri2026reference}); ISO/IEC = International Organization for Standardization/International Electrotechnical Commission (42001, \citeyear{iso42001}; 42005, \citeyear{iso42005}). ARC = Agentic Risk \& Capability Framework \citep{khoo2025}; OWASP = Open Worldwide Application Security Project \citep{owasp2025}. Delegation research refers to \citet{feng2025}, \citet{zheng2026}, and \citet{arora2026}; MeRGE refers to \citet{fetzer2026}.}
\endgroup

One failure can span domains: goal hijacking may exploit an authorization defect, expose private data, and cause financial loss. Retain that causal chain as one linked scenario rather than several independent events. A primary domain supports ownership; dependencies remain distinguishable from impacts, and concerns that fit poorly remain visible. Resource cost falls under resilience when it threatens continued service and under Mission \& Value when it defeats the economic objective.

\subsection{Assurance: Specify What Must Be Demonstrated}
Assurance specifies evaluations needed before deployment or material change. This is AI-GRACE's operational distinction; broader assurance guidance covers trustworthiness across development and deployment \citep{dsit2024}. An assurance requirement states the claim, test method, acceptance criterion, population, workload, version, and treatment of uncertainty.

Tests address the assembled system and its intended controls, covering task quality, agent behavior, tool use, security, privacy, human impact, resilience, scale, and economics as relevant (\citealp[MS-2.1.1--MS-2.1.2, MS-2.3.4, MS-2.5.1, MS-2.12.3]{cri2026reference}; \citealp[MEASURE 2.1, 2.3--2.7, 2.10--2.11]{nist2023}). Tests must exercise the actual integration path. An aggregate accuracy score can obscure a small but unacceptable set of errors.

Changes in actions or authority require corresponding evaluation, not a universal test suite for each RAIL. Failed assurance can lead to redesign, stronger controls, reduced scope, or a decision not to deploy.

\subsection{Controls: Constrain and Sustain Operation}
ISO/IEC 42001 and NIST use controls broadly, including organizational policies and procedures (\citealp[clause 3.21]{iso42001}; \citealp[GOVERN 1.4]{nist2023}). AI-GRACE's Controls function focuses on technical mechanisms that enforce boundaries and sustain service. For example, a policy restricts access to protected data. Then identity-based authorization enforces that restriction when the agent requests data or invokes a tool.

Controls include guardrails, trusted tool interfaces, identity and authorization, approval workflows, action limits, resource budgets, isolation, failover, and revocation. A guardrail inspects inputs, outputs, or tool calls and blocks, modifies, or escalates activity against a policy. Its detection may be imperfect. So, consequential prohibitions should, where feasible, also be enforced independently of the agent's interpretation.

Control placement follows the execution path. Input inspection addresses untrusted content, orchestration limits trajectories and budgets, data services enforce access scope, and execution services constrain consequential actions. Each control specifies its condition, location, required identity and state, dependencies, owner, failure behavior, and evidence. Teams must justify each additional control layer's latency, cost, complexity, and failure modes against the service objective.

\subsection{Evidence: Retain the Basis for Reliance}
Evidence connects deployment decisions to evaluations and operational outcomes. It includes risk decisions, versions, control tests, approvals, observable actions, and business and service results, linked to the requirements they substantiate (\citealp[MS-2.4.1--MS-2.4.3]{cri2026reference}; \citealp[MEASURE 2.4; MANAGE 4.1]{nist2023}). For a consequential action, the record links the principal and agent to the request and its authorization under a specified mandate and policy version. It retains relevant provenance, distinguishes what was attempted from what was committed, and records the outcome.

The evidence design distinguishes an agent's account of an action from the authoritative execution record. Collection remains proportionate to purpose, with access controls, integrity protection, retention periods, and justified sampling. AI-GRACE does not require hidden model reasoning or indiscriminate retention of sensitive data.

Detailed interaction capture can aid diagnosis: Langfuse supports tracing prompts, responses, tool and retrieval steps, timing, and metadata \citep{langfuse}. A team adopting such capture must qualify its privacy, access, and retention properties. Critical action records may need different availability and retention from aggregate quality monitoring.

\subsection{Derive and Qualify the Enabling Capabilities}
A capability requirement specifies what must hold, where, and how adequacy will be demonstrated. The derivation procedure is:
\begin{enumerate}
\item Select an objective, obligation, or material risk and state the required outcome.
\item Specify what must be demonstrated before reliance, constrained during operation, and retained as evidence.
\item Identify the technical capabilities and supporting services needed.
\item Define acceptance criteria, scope, interfaces, dependencies, owners, and architectural placement.
\item Assess candidate implementations and record gaps or justified alternatives.
\end{enumerate}

Requirements can originate directly in objectives or obligations. One service can support several A/C/E requirements, and one requirement can depend on several services. Model serving, storage, connectivity, capacity, isolation, and recovery are enabling capabilities whose required properties follow from the use case.

For each candidate implementation, record one of four fit states:
\begin{description}[style=nextline]
\item[Sufficient.] Evidence demonstrates adequacy within the stated scope, version, workload, and dependencies.
\item[Partially sufficient.] Some required properties are demonstrated, but identified limitations remain.
\item[Absent.] No implementation of the required capability is available.
\item[Not yet demonstrated.] A candidate exists, but has not been evaluated.
\end{description}

Owning an evaluation tool does not demonstrate that it can evaluate this use case, or that this agent passed its evaluations. Prior qualification applies only where its conditions still hold. The gap record names the unmet requirement, responsible owner, closure evidence, dependencies, alternatives, and effect on deployment. Missing legal permission or accountability remains a separate constraint.

\subsection{Realize the Requirements in a Logical Architecture}
The logical design places capabilities along the workflow: principals, agent and model services, data access, consequential actions, enforcement points, evidence paths, and infrastructure. Each capability has a location or an unresolved gap, and each material control links to its assessment basis. ISO/IEC 42001's resource and technical documentation provisions provide relevant foundations \citep[Annex B.4 and B.6.2.7]{iso42001}.

Review whether identity and access scope survive delegation, required approval precedes execution, retries avoid duplicate or inconsistent effects, dependencies support recovery, and evidence can be reconciled. Assess the cost and latency of the complete design. An approved target architecture supports engineering. Deployment readiness requires evidence that the implemented configuration satisfies it. Product selection, bills of materials, and code follow from that design.

\subsection{Determine Authority and Summarize It With RAIL}
Deployment requires demonstrated technical support, permission under applicable law, contract, and organizational policy, and an accountable decision accepting residual risk and the operating arrangements. None of these conditions establishes the others. A requested operating envelope does not establish authorization.

The decision covers the whole operating envelope, including concurrent actions, their sequence, aggregate exposure, and dependencies. If no acceptable envelope remains, deployment is not authorized. An organization may authorize less than it could technically support.

The Agent Operating Envelope defines permitted and prohibited actions and any required approvals within a specified scope of accounts, data, tools, and counterparties. It sets operational limits and mandate duration, bounds resource use, and specifies escalation and revocation conditions. Organizational deployment authorization and any required consent from affected parties remain distinct. Recommendations can have consequential effects even without execution.

Table~\ref{tab:rail} summarizes authorized independence. The envelope remains authoritative; RAIL is not a maturity rating, capability score, or measure of risk.

\begingroup\normalsize\singlespacing
\setlength{\LTleft}{0pt}
\setlength{\LTright}{\fill}
\begin{longtable}{P{1cm}P{3.0cm}P{\dimexpr(\textwidth-6\tabcolsep-1cm-3.0cm)/1\relax}}
\caption{Risk-Aligned Independence Levels}
\label{tab:rail}\\
\toprule \colhead{Level} & \colhead{Label} & \colhead{Authorized behavior} \\
\midrule\endfirsthead
\toprule \colhead{Level} & \colhead{Label} & \colhead{Authorized behavior} \\
\midrule\endhead
0 & Assist & Retrieve, summarize, or present information within scope; no mandate to recommend a course of action or execute consequential external actions. \\
1 & Recommend & Analyze and recommend; a human decides and performs consequential external actions. \\
2 & Act with Approval & Each consequential external action requires explicit human approval. Scoped retrieval, reasoning, and preparation do not require approval at every step. \\
3 & Bounded Autonomous & Execute a defined class of consequential actions within an approved mandate without individual approval; escalate at defined boundaries. \\
4 & High-Authority Autonomous & Coordinate broader approved tasks or mandates under supervision rather than routine approval of individual consequential actions. All activity remains bounded. \\
\bottomrule
\end{longtable}
\endgroup

Mixed workflows retain conditions for each action. RAIL classification follows the approved behavior and scope of each action. A higher level does not permit an agent to expand its own privileges. If the requested scope cannot be supported, the assessment records the gap and available alternatives.

\subsection{Reassess the Use Case and Reuse Qualified Patterns}
Profiles organize context through core, jurisdiction, industry, organization, and use case, consistent with NIST's contextual tailoring \citep[section 6]{nist2023}. For example, industry profiles can include financial services, healthcare, higher education, defense, and critical infrastructure. Their baseline Governance and Risk inputs would cover obligation references, governance expectations, affected parties, and recurring scenarios. Developing and evaluating these profiles remains future work.

Within a deployment, changed behavior, cost, obligations, data, models, tools, dependencies, scale, or authority triggers targeted reassessment. A protective response may narrow or suspend activity. Expansion requires a new decision. An absence of incidents is meaningful only where detection and evidence collection could reveal failures.

Across use cases, teams can reuse applicable assessments, qualified capabilities, architectures, and evidence after checking scope and currency. Shared dependencies, capacity contention, correlated failures, and combined permissions may require assessment across deployments. Reuse does not transfer RAIL or authorization.

\section{Illustrative Application: A Personal Banking Assistant}
\label{sec:retail}
A fictional banking group proposes a personal banking assistant for its U.S. national bank and EU credit institution. Conventional banking infrastructure exists, but no assistant is deployed or authorized. The service would explain accounts and spending, review bills, and arrange client-approved payments.

The scenario assumes full applicability of the EU AI Act and DORA to the relevant entities and activities after the applicable dates. Obligations, starting capabilities, risk judgments, and acceptance conditions are illustrative assessment inputs. Tables~\ref{tab:bank-governance}--\ref{tab:fit} trace the proposal through governance, risk, A/C/E requirements, and capability qualification.

\subsection{Governance: Establish the Service and Applicable Obligations}
The service owner seeks correct task completion with less client effort and sustainable total cost, compared with conventional digital banking and human assistance. ``Explain my electricity bill and help me pay it'' must end with a correct explanation and, if the client proceeds, the payment they approved. Pending payments, unresolved conversations, and repeat calls to correct errors do not count as successful completion.

Scope includes authenticated clients, authorized accounts, approved bank information, bills, and existing payment services. Creditworthiness decisions, credit scoring, product eligibility, and investment trading are excluded. The EU entity is assumed to provide the system under its own name and use it, requiring assessment of provider and deployer responsibilities; intended purpose determines classification \citep[Articles 3 and 6]{euaiact2024}. This assumption does not make every banking assistant high-risk.

\begingroup\normalsize\singlespacing
\setlength{\LTleft}{0pt}
\setlength{\LTright}{\fill}
\begin{longtable}{P{0.65cm}P{5.9cm}P{\dimexpr(\textwidth-6\tabcolsep-0.65cm-5.9cm)/1\relax}}
\caption{Governance Inputs and Required Outcomes}
\label{tab:bank-governance}\\
\toprule \colhead{ID} & \colhead{Source and applicability} & \colhead{Required outcome} \\
\midrule\endfirsthead
\toprule \colhead{ID} & \colhead{Source and applicability} & \colhead{Required outcome} \\
\midrule\endhead
G1 & Bank service charter: organizational objectives and commitments. & Correct completion, accessible human recourse, fewer interactions, and sustainable total cost, including failures and rework. \\
G2 & EU AI Act: Article 5(1)(a)--(b) prohibitions; Article 50(1), (5) AI interaction notice; Article 50(2) marking of generated content; recital 27's nonbinding principles \citep{euaiact2024}. & Prevent prohibited manipulation or exploitation meeting the Act's conditions; identify AI interaction. Record Article 50(2) applicability and any required marking of generated content. The bank adopts oversight, safety, fairness, and accountability as service requirements, including for vulnerable clients. \\
G3 & GDPR Article 5: lawful processing, purpose limitation, minimization, security, accountability \citep{gdpr2016}. & Constrain account access, retrieval, memory, model disclosure, and telemetry to the approved purpose; demonstrate these boundaries. \\
G4 & U.S. national bank: Gramm--Leach--Bliley Act safeguards through 12 CFR part 30, Appendix B, II--III \citep{occsecurity}. & Extend customer information safeguards to the agent, tools, and suppliers: threat assessment, access restrictions, protection, testing, and incident response. \\
G5 & EU credit institution: DORA Articles 8, 11, 17--19, 24--25, 28--30 \citep{dora2022}. & Map dependencies; test continuity and recovery; investigate and report qualifying ICT incidents through bank processes; assess suppliers and contracts. \\
G6 & U.S. Regulation E, 12 CFR 1005.11; national implementation of PSD2 Articles 64, 97--98 \citep{rege,psd22015}. & Preserve consent and applicable authentication, including transaction linking where required; reconstruct errors and disputes. The chosen design requires client approval of each payment alongside existing bank checks. \\
\bottomrule
\end{longtable}
\apanote{G identifies Governance inputs. EU = European Union; GDPR = General Data Protection Regulation; CFR = Code of Federal Regulations; DORA = Digital Operational Resilience Act; ICT = information and communication technology; PSD2 = Second Payment Services Directive.}
\endgroup

Compliance owns the applicability record, including national implementation and the distinction between binding provisions and adopted principles. Service, privacy, security, payments, and operations owners translate these obligations into technical requirements. Any required marking under Article 50(2) enters A1 tests, C1 output processing, and E1's configuration and evaluation records (Table~\ref{tab:bank-ace}).

\subsection{Risk: Identify the Failure Paths That Matter}
Table~\ref{tab:risks} covers ordinary failures and malicious influence. For this greenfield workflow, release decisions depend on consequences and acceptance criteria.

\begingroup\normalsize\singlespacing
\setlength{\LTleft}{0pt}
\setlength{\LTright}{\fill}
\begin{longtable}{P{3.0cm}P{\dimexpr(\textwidth-6\tabcolsep-3.0cm)/2\relax}P{\dimexpr(\textwidth-6\tabcolsep-3.0cm)/2\relax}}
\caption{Risks Linked to Governance Inputs and Treatment Decisions}
\label{tab:risks}\\
\toprule \colhead{Risk / governance link} & \colhead{Failure and consequence} & \colhead{Required property and owner} \\
\midrule\endfirsthead
\toprule \colhead{Risk / governance link} & \colhead{Failure and consequence} & \colhead{Required property and owner} \\
\midrule\endhead
R1: AI Behavior \& Authority; G1, G6 & Invented balances, misread bills, wrong tools, or false completion produce incorrect advice or payment. & Ground answers and verify outcomes; service/payments owners require critical task and action tests. \\
R2: Mission \& Value; G1 & Repeated calls and unresolved tasks increase cost and client effort. & Measure quality and total cost per resolved task; service owner rejects failure to meet value criteria. \\
R3: Cybersecurity; G2, G4, G5 & Injected bill content or compromised tools redirect the agent toward theft or altered payments. & Security owner requires adversarial tests, tool assessment, and independent enforcement. \\
R4: Privacy; G3, G4 & Retrieval, memory, model requests, or traces expose another client's data or retain excess information. & Privacy owner requires client/entity isolation, limited collection, and access/leakage tests across data paths. \\
R5: Operational Resilience; G5, G6 & Model/tool failure stalls requests; retries duplicate payments or lose their status. & Operations owner requires consistent recovery, reconciliation, fallback, and workload/failure tests. \\
R6: Safety \& Human Impact; G1, G2 & Manipulative guidance, harmful reliance, unequal service, or inaccessible assistance harms clients. & Service owner requires tests by affected group, safe responses, and recourse; critical harm blocks release. \\
R7: Legal, Regulatory \& Policy; G2--G6 & Missing disclosures, bypassed consent, mishandled incidents, or crossed service/entity boundaries breach obligations. & Compliance records permissibility; deployment authority reviews remaining gaps and evidence of enforcement. \\
\bottomrule
\end{longtable}
\apanote{R identifies risk scenarios; G identifies the linked Governance inputs in Table~\ref{tab:bank-governance}.}
\endgroup

Malicious instructions in a bill can redirect the agent, exploit an access-control weakness, and expose account data or cause an incorrect payment; these consequences belong to one linked scenario. R1 can cause an incorrect payment without an attacker. An accurate answer may still be manipulative or unsuitable under R6, so quality, safety, and security require different demonstrations.

\subsection{Assurance: Derive the Technical Evaluation Capabilities}
Table~\ref{tab:bank-ace} derives the technical specification. R7 spans these chains through the applicable obligations. The design uses the Model Context Protocol (MCP) for selected tool connections.

\begingroup\normalsize\singlespacing
\setlength{\LTleft}{0pt}
\setlength{\LTright}{\fill}
\begin{longtable}{P{1.55cm}P{\dimexpr(\textwidth-8\tabcolsep-1.55cm)/3\relax}P{\dimexpr(\textwidth-8\tabcolsep-1.55cm)/3\relax}P{\dimexpr(\textwidth-8\tabcolsep-1.55cm)/3\relax}}
\caption{Linked Assurance, Control, and Evidence Capabilities}
\label{tab:bank-ace}\\
\toprule \colhead{Risk basis} & \colhead{Assurance: before release or change} & \colhead{Controls: runtime enforcement} & \colhead{Evidence: tests and outcomes} \\
\midrule\endfirsthead
\toprule \colhead{Risk basis} & \colhead{Assurance: before release or change} & \colhead{Controls: runtime enforcement} & \colhead{Evidence: tests and outcomes} \\
\midrule\endhead
R1, R2, R6; G1--G2 & \textbf{A1 Agent evaluations.} Banking tasks: correctness, tool choice, completion, harmful advice, group differences, handoff, effort, cost. & \textbf{C1 Quality and safety guardrails.} Grounding, output checks, scope limits, AI notice, human recourse. & \textbf{E1 Agent tracing and outcome review.} Model/tool spans, sources, outputs, resolution, correction, handoff, cost; linked evaluation versions. \\
\addlinespace[0.35em]
R3, R6; G2, G4--G5 & \textbf{A2 Red teaming and MCP scanning.} Model/agent attacks, injected content, tool descriptions, servers, permissions, dependencies. & \textbf{C2 Runtime inspection.} Inspect input, output, and tool results; permit only approved tool and server versions; contain or escalate suspicious activity. & \textbf{E2 Security and safety findings.} Test cases, scanner coverage, findings and disposition, guardrail events, policy versions, incidents. \\
\addlinespace[0.35em]
R4; G3--G4 & \textbf{A3 Access and privacy tests.} Client/agent/entity permissions, session isolation, leakage, telemetry minimization. & \textbf{C3 Identity-based enforcement.} Scoped credentials, authorization at tool/data boundaries, isolated memory, protected/redacted telemetry. & \textbf{E3 Access decisions.} Principal, resource, policy, permit/deny result, retention/access configuration; exclude unnecessary payloads. \\
\addlinespace[0.35em]
R1, R3, R5; G6 & \textbf{A4 Payment integration tests.} Approval tampering, revocation, replay, timeout, reconstruction through execution. & \textbf{C4 Transaction enforcement.} Approval bound to exact details; independent gate, durable duplicate suppression, reconciliation. & \textbf{E4 Payment evidence.} Proposal, client approval, authorization, committed transaction, reconciled status; linked identifiers. \\
\addlinespace[0.35em]
R5, R2; G1, G5 & \textbf{A5 Resilience and load tests.} Outages, capacity, recovery, telemetry loss, incident routing. & \textbf{C5 Operational constraints.} Timeouts, retry/step/cost limits, circuit breakers, recoverable state, payment holds, fallback. & \textbf{E5 System observability.} Correlated logs, traces, metrics, dependency health, service objectives, incidents, recovery results. \\
\bottomrule
\end{longtable}
\apanote{A = Assurance; C = Controls; E = Evidence; MCP = Model Context Protocol. G and R refer to the linked inputs and risks in Tables~\ref{tab:bank-governance} and~\ref{tab:risks}.}
\endgroup

Domain experts review A1 tasks against conventional service: resolution quality must not degrade, interactions must decrease, and total cost per resolved request must not increase, including failed attempts and correction. Critical cases cover wrong accounts, payees, or amounts; false completion; harmful guidance; and failed handoff. Each must meet its expected outcome, with results reviewed by task and affected client group.

A2 uses red teaming to test how the model and assembled agent respond to attacks. MCP scanning checks selected servers and their metadata, covering configurations and dependencies as well as permissions. MCP guidance identifies risks involving authorization and trust boundaries \citep{mcpsecurity2025}. Scans cover only their stated checks and require targeted execution tests; safe model responses alone do not establish safe tool use.

Under A3--A5, a changed payee after approval must be rejected, and a timeout after commitment must trigger reconciliation rather than another payment. Release requires a qualified evaluation process and evidence that the actual configuration passed, with versions, coverage, failures, and residual uncertainty recorded.

\subsection{Controls: Place Enforcement on the Runtime Paths}
C1/C2 inspect client requests, retrieved content, tool results, and responses. Detection remains imperfect; staying on topic does not establish correctness. C3 independently enforces access restrictions at the gateway and downstream service, using verifiable client and agent identities, entity context, account scope, and the requested data and action. It also isolates sessions and limits model context to what the task requires.

For C4, the client reviews canonical details in the trusted bank interface. Approval binds principal, entity, account, payee, amount, currency, execution date, and validity conditions; the execution service checks the record and current policy before commitment. Changed details require new approval, and the agent has no bypass credential.

C5 holds ambiguous payments for reconciliation, caps repeated work, and routes unresolved requests to human service. Loss of a required payment control or durable action record blocks new commitments.

\subsection{Evidence: Connect Agent Behavior to Service and System Outcomes}
The collector correlates agent spans, control decisions, system logs, metrics, and backend outcomes. A ``payment sent'' trace must join to the authoritative payment result before it supports completion. E1--E5 also retain evaluations, versions, approvals, and release decisions, subject to access, redaction, retention, and integrity controls.

Quality sampling cannot discard required payment records. Report unresolved outcomes explicitly. Token estimates alone do not establish total service cost. Production traces can inform renewed evaluations without replacing assurance before release.

\subsection{Architecture: Assemble the Agent and Its Enabling Services}
\begin{figure}[p]
\caption{Target Architecture for the Greenfield Banking Assistant}
\label{fig:architecture}
\raggedright
\begingroup
\singlespacing\hyphenpenalty=10000\exhyphenpenalty=10000
\begin{tikzpicture}[
  font=\fontsize{9}{11}\selectfont\sffamily,
  box/.style={draw,line width=0.6pt,rounded corners=2pt,align=center,inner sep=5pt,text width=3.05cm,minimum height=2.05cm},
  guard/.style={box,fill=black!6},
  arr/.style={-{Latex[length=1.7mm]},line width=0.7pt},
  both/.style={{Latex[length=1.5mm]}-{Latex[length=1.5mm]},line width=0.7pt},
  ev/.style={-{Latex[length=1.5mm]},dashed,line width=0.6pt},
  tag/.style={font=\fontsize{9}{11}\selectfont\sffamily,fill=white,inner sep=2pt,align=center}
]

\node[box] (channel) at (0,0) {\textbf{Bank app/client}\\Authenticated session\\AI notice (C1)\\Human service access};
\node[guard] (content) at (4,0) {\textbf{Guardrails (C1/C2)}\\Input / output checks\\Safety and quality\\Injection inspection};
\node[box] (agent) at (8,0) {\textbf{Agent harness}\\Planner / task state\\Tool router / MCP\\Step/cost limits (C5)};
\node[guard] (modelgate) at (12,0) {\textbf{Model gateway}\\C2 inspection\\Model selection\\Versioned routing};
\node[box] (memory) at (0,-2.8) {\textbf{Session store}\\Client/entity isolation\\Scoped memory (C3)};
\node[guard] (identity) at (4,-2.8) {\textbf{Identity / policy}\\Client / agent / entity\\Scoped credentials\\C3 authorization};
\node[guard] (toolgate) at (8,-2.8) {\textbf{MCP / API gateway}\\C2 traffic checks\\C3 authorization\\Tool allowlist};
\node[box] (model) at (12,-2.8) {\textbf{Model service}\\Versioned model\\Scoped context\\Serving capacity};
\draw[both] (channel.east)--(content.west);
\draw[both] (content.east)--(agent.west);
\draw[both] (agent.east)--(modelgate.west);
\draw[both] (modelgate)--(model);
\draw[both] (agent.south west)--(6,-1.4)--(0,-1.4)--(memory.north);
\draw[both] (agent)--(toolgate);
\draw[arr] (identity)--(toolgate);

\node[guard] (approval) at (0,-5.6) {\textbf{Approval UI (C4)}\\Exact transaction\\Trusted client review\\Bound approval record};
\node[guard] (execution) at (4,-5.6) {\textbf{Payment gate}\\C3 / C4 checks\\Approval and policy\\No bypass};
\node[box] (tools) at (8,-5.6) {\textbf{Bank tool adapters}\\Account / bill tools\\Knowledge retrieval\\Payment tools};
\node[box] (accounts) at (12,-5.6) {\textbf{Account / bill APIs}\\Client accounts/bills\\C3 account scope\\Authoritative records};
\node[box] (payments) at (4,-8.4) {\textbf{Payment services}\\Bank checks (C4)\\Execute / deduplicate\\Reconciled status (E4)};
\node[box] (knowledge) at (12,-8.4) {\textbf{Bank knowledge}\\Bank policies / FAQs\\Retrieval index\\Source/version};
\draw[both] (toolgate)--(tools);
\draw[both] (tools)--(accounts);
\draw[both] (tools.south east)--(10,-7.0)--(10,-8.4)--(knowledge.west);
\draw[both] (tools.west)--(execution.east);
\draw[both] (approval.east)--(execution.west);
\draw[arr] (identity)--(execution);
\draw[both] (execution)--(payments);
\draw[both] (channel.west)--(-2.0,0)--(-2.0,-5.6)--(approval.west);

\node[draw,densely dotted,rounded corners=3pt,inner sep=9pt,
  fit=(channel)(modelgate)(approval)(payments)(knowledge)] (runtime) {};
\node[tag,anchor=south] at (6,1.48) {\textbf{Personal banking assistant: configuration for each banking entity}};

\node[box,text width=6.3cm,minimum height=2.5cm] (assurance) at (1.35,-11.9) {\textbf{Assurance environment (A1--A5)}\\Agent evaluations; model/agent red teams\\MCP server / tool / dependency scanning\\Access, payment, load, and recovery tests};
\node[box,minimum height=2.5cm] (collector) at (8,-11.9) {\textbf{Telemetry collector}\\Continuous agent spans\\Logs / metrics / events\\Correlation; redaction};
\node[box,minimum height=2.5cm] (evidence) at (12,-11.9) {\textbf{Evidence / review}\\E1--E5 records\\Dashboards / alerts\\Tests and decisions\\Incidents / outcomes};
\draw[ev] (assurance.north)--node[tag,left]{exercise assembled\\configuration}(runtime.south -| assurance.north);
\draw[ev] (runtime.south -| collector.north)--node[tag,right]{instrumentation\\from runtime components}(collector.north);
\draw[ev] (assurance)--node[tag,above]{test results}(collector);
\draw[arr] (collector)--(evidence);
\node[box,text width=14.7cm,minimum height=1.35cm] (platform) at (6,-14.4) {\textbf{Supporting infrastructure and operations (A5/C5)}\\Compute, network/egress isolation, storage, secrets, queues, backup/recovery, dependency and supplier management};
\draw[ev] (platform.east)--(14.0,-14.4)--(14.0,-4.2)--(runtime.east);
\end{tikzpicture}
\endgroup
\apanote{Solid arrows show service and tool paths; dashed arrows connect assurance, telemetry, and infrastructure. A = Assurance; C = Controls; E = Evidence; MCP = Model Context Protocol; API = application programming interface; UI = user interface; FAQ = frequently asked question. A1--A5, C1--C5, and E1--E5 refer to Table~\ref{tab:bank-ace}.}
\end{figure}
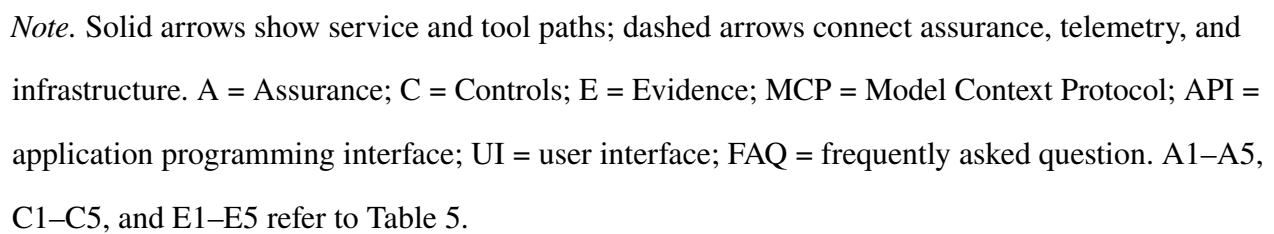

Figure~\ref{fig:architecture} places the capabilities in the agent's request, model, data, and execution paths. The harness maintains task state, calls the model, selects tools, and prepares responses or actions; account and bill APIs remain authoritative. Assurance exercises these interfaces, while runtime instrumentation feeds observability and retained evidence. EU and U.S. deployments use their own policy, data, and supplier configurations.
\clearpage

\subsection{Capability Qualification and the Deployment Decision}
Table~\ref{tab:fit} assesses assumed bank capabilities against the new assistant's requirements. Prior qualification of a banking service grants no assistant access by itself.

\begingroup\normalsize\singlespacing
\setlength{\LTleft}{0pt}
\setlength{\LTright}{\fill}
\begin{longtable}{P{4.5cm}P{3.25cm}P{\dimexpr(\textwidth-6\tabcolsep-4.5cm-3.25cm)/1\relax}}
\caption{Capability Fit and Required Delivery Work}
\label{tab:fit}\\
\toprule \colhead{Candidate capability} & \colhead{Illustrative fit} & \colhead{Required work and owner} \\
\midrule\endfirsthead
\toprule \colhead{Candidate capability} & \colhead{Illustrative fit} & \colhead{Required work and owner} \\
\midrule\endhead
Bank identity, read APIs, payment services & Existing bank services: sufficient. Assistant integration: not yet demonstrated. & Architecture/security owners verify the new agent and delegation paths. \\
Agent evaluations, red teaming, MCP scanning, access/privacy tests, payment integration tests (A1--A4) & Absent for this use case & Evaluation/security owners establish tasks, attacks, assessments, and acceptance evidence. \\
Runtime guardrails, agent access and approval enforcement (C1--C4) & Absent at agent boundary & Engineering/security/payments owners add inspection, scoped authorization, approval binding, and execution mediation; qualify through A1--A4. \\
Central logging and bank transaction records (E1--E5) & Partially sufficient & Observability/evidence owners add continuous tracing, correlated decisions, protected retention, outcome links, and incident integration; test reconstruction and collection failure. \\
Hosting, model serving, recovery, suppliers (A5/C5) & Not yet demonstrated for design & Platform/operations owners establish capacity, dependency, recovery, and supplier evidence. \\
\bottomrule
\end{longtable}
\apanote{A = Assurance; C = Controls; E = Evidence; API = application programming interface; MCP = Model Context Protocol. Identifiers refer to Table~\ref{tab:bank-ace}.}
\endgroup

The current decision authorizes engineering and evaluation, with no assistant deployment yet authorized. Release requires technical gap closure, acceptable value and residual risk, and recorded legal/policy permissibility for each entity.

If qualified and authorized, retrieval and preparation can proceed within account scope while each payment requires client approval: RAIL~2. The envelope fixes tools, payees, existing bank transaction limits, approval expiry, resource budgets, entity/data boundaries, escalation, and revocation. Scoped reads and model calls need no separate approval.

\subsection{Reassessment and Reuse When the Service Changes}
A new bill source reopens provenance, MCP/tool configuration, injection, privacy, supplier, and quality checks under A1--A3/A5. Approval and payment evidence patterns can be reused if the execution path is unchanged; the source remains outside the envelope until gaps close. Rising cost per resolved task reopens R2, C5 budgets, and A1 evaluation without necessarily changing payment authority.

Delegated wealth activity requires an updated charter and applicable investment permissions, considering U.S. fiduciary interpretation and robo-adviser staff guidance where relevant \citep{sec2019,sec2017}. Excessive equities trading volume could arise from malicious influence, poor reasoning, a misaligned objective, or an uneconomic strategy. Assurance adds mandate and market stress tests; controls add instrument, exposure, turnover, validity, and cost limits; evidence links positions, client costs, exceptions, and supervision. Historical simulation does not establish future performance, and transaction revenue does not establish client value.

A bounded investment mandate may support RAIL~3 after separate authorization. Broader coordination may require RAIL~4 and combined constraints (investment activity must not consume liquidity reserved for an approved bill). Identity, tracing, and execution patterns may be reusable, but mandates, tests, and authorization are not inherited.

\section{Discussion}
\subsection{Value, Proportionality, and Reuse}
AI-GRACE keeps service value visible when selecting technical capabilities, consistent with NIST and CRI's attention to business value and resources \citep{nist2023,cri2026guide}. Teams must compare the cost and effects of controls with the intended outcome while meeting mandatory obligations. Reuse may reduce repeated work, but maintaining qualifications, dependencies, and shared services also creates cost.

\subsection{Limitations}
AI-GRACE has not been empirically validated. The purposive source selection and fictional application support its design rationale, but do not show whether the method improves deployment decisions. The coverage of the seven risk domains and the reliability of capability fit judgments require independent evaluation.

Assessment quality depends on the available evidence and the assessors' judgment. Incomplete evidence, missed risks, or implementation defects can leave a deployment exposed despite a completed assessment. Shared platforms and interacting agents may also create effects beyond an individual assessment's scope.

Initial studies should compare AI-GRACE with established organizational practice informed by relevant standards, using matched deployment cases and independent review of missed requirements, unsupported capability judgments, architecture consistency, and assessment effort. Field studies should examine implementation rework, operating outcomes, and reuse. A justified decision to narrow or reject deployment should count as a useful outcome. Further work should develop and evaluate industry profiles and test assessments across interacting deployments.

\begingroup\widowpenalty=10000
\section{Conclusion}
Organizations deploying agentic AI commit resources and accept responsibility for systems that can affect people, data, and operations. They need to establish whether the proposed use case can deliver value, meet its obligations, and operate within enforceable boundaries. Authorizing deployment without a structured assessment of those conditions would place the organization and its stakeholders at risk.

AI-GRACE connects those responsibilities to the technical capabilities and architecture an organization must put in place. It identifies what the organization can already support, what remains unresolved, and the evidence needed to justify deployment. Organizations need AI-GRACE or an equivalent method to connect governance commitments to implementation and remain accountable for the scope they authorize. The purpose is to pursue risk-adjusted value while retaining a basis for reassessment and reuse.

\endgroup

\clearpage

\begingroup\normalsize\doublespacing
\setlength{\bibsep}{0pt}
\setlength{\bibhang}{0.5in}

\endgroup

\appendix
\renewcommand{\thetable}{\thesection\arabic{table}}
\clearpage
\refstepcounter{section}\label{app:record}
\setcounter{table}{0}
\section*{Appendix \thesection\\Minimum Deployment Record}
Table~\ref{tab:record} specifies linked information, not separate mandatory documents. Existing organizational systems can hold these records, with owners, versions, approval status, and review dates or triggers.

\begingroup\normalsize\singlespacing
\setlength{\LTleft}{0pt}
\setlength{\LTright}{\fill}
\begin{longtable}{P{3.2cm}P{12.0cm}}
\caption{Minimum Content of an AI-GRACE Assessment}\label{tab:record}\\
\toprule \colhead{Record} & \colhead{Required content} \\
\midrule\endfirsthead
\toprule \colhead{Record} & \colhead{Required content (continued)} \\
\midrule\endhead
Use case charter & Objectives, impacts, comparator, acceptance conditions, affected parties, organizational and affected-party roles, workflow, data, tools, actions, exclusions. \\
Obligation register & Source and locator; legal, contractual, policy, or strategic status; applicability, jurisdiction, owner, requirements, conflicts, unresolved decisions. \\
Risk register & Scenario, causes, primary/linked domains, affected objectives and parties, likelihood, consequences, assumptions, uncertainty, treatment, residual risk, decision owner. \\
Capability requirement & Identifier, linked objective/obligation/risk, A/C/E function, scope, acceptance criterion, evidence method, placement, interfaces, dependencies, owner. \\
Capability fit & Candidate implementation/version, fit state, supporting evidence, qualification scope, limitations, required changes or validation. \\
Logical design & Agent/services, principals, data and execution paths, enforcement, evaluations, evidence, infrastructure, requirement links, gaps. \\
Operating envelope & Allowed and prohibited activity, required approvals; accounts, data, tools, counterparties, limits, duration, resources, escalation, revocation. \\
Decision and gaps & Desired envelope/RAIL, technical support, permissibility, actual authorization or refusal, accountable decision, blockers, owners, closure evidence, review conditions. \\
Reuse record & Source pattern, assumptions, qualification scope, applicable requirements/evidence, changed elements, reassessment, expiry or invalidation triggers. \\
\bottomrule
\end{longtable}
\endgroup
\apanote{A = Assurance; C = Controls; E = Evidence; RAIL = Risk-Aligned Independence Levels.}

\clearpage
\refstepcounter{section}\label{app:authorship}
\section*{Appendix \thesection\\Evidence and Authorship Statement}
\grace{} originated in the authors' professional observations of the challenges organizations face in translating AI objectives and obligations into technical implementation. The framework was conceived and developed as a potential solution to those practical deployment challenges.

AI tools were used as research and editorial assistants to accelerate literature discovery for manual review, manuscript editing and clarification, and adversarial refinement of the framework and its illustrative application. This included challenging assumptions, identifying gaps, and testing the clarity and consistency of the reasoning. The authors directed this process and retained responsibility for the framework's design and substantive decisions. They stand behind the work and take responsibility for its claims.

The manuscript presents a proposed framework and a fictional retail-banking application. The demonstration contains no real client data, and this version reports no organizational experiment or empirical validation of the framework.

\clearpage

\end{document}